\documentclass[letterpaper,10pt, conference]{ieeeconf}  

\usepackage{tikz}
\usetikzlibrary{arrows}
\usetikzlibrary{shapes.geometric, arrows}
\usetikzlibrary{positioning,fit,shapes}
\usepackage{verbatim}
\usepackage{graphics} 
\usepackage{epsfig} 
\usepackage{mathptmx} 
\usepackage{times} 
\usepackage{amsmath} 
\usepackage{amssymb}  
\usepackage{cite}
\usepackage{float}
\usepackage{placeins}
\usepackage{textcomp}
\usepackage{graphicx}
\usepackage{color}
\usepackage{epstopdf}
\graphicspath{{images/}}
\usepackage[linesnumbered,ruled,vlined]{algorithm2e}
\usepackage{algorithmic}
\usepackage{graphicx}
\graphicspath{ {images/} }

\title{\LARGE \bf Securing a UAV Using Individual Characteristics From an EEG Signal }
  \author{Ashutosh~Singandhupe,
	Hung~Manh~La,~\IEEEmembership{Senior Member,~IEEE,}
	David~Feil-Seifer,
	Pei~Huang,
	Linke~Guo,
	and Ming~Li
\thanks{Ashutosh Singandhupe and Dr. Hung La are with the Advanced Robotics and Automation (ARA) Lab, 
 Department of Computer Science and Engineering, University of Nevada, Reno, NV, 89557, USA}
\thanks{Dr. David Feil-Seifer and Dr. Ming Li  are with the Department of Computer Science and Engineering, University of Nevada, Reno, NV, 89557, USA}
\thanks{ Pei Huang and Dr. Linke Guo  are with Department of Electrical and Computer Engineering
Binghamton University, State University of New York, Binghamton, NY, 13902, USA.}
\thanks{\emph{Corresponding author:} Hung Manh La (e-mail: hla@unr.edu).}%
}

\begin{document}
\vspace{-40pt}
\maketitle
\begin{abstract}
   Unmanned aerial vehicles (UAVs) have gained much attention in recent years for both commercial and military applications. The progress in this field has gained much popularity and the research has encompassed various fields of scientific domain. Cyber securing a UAV communication has been one of the active research field since the attack on Predator UAV video stream hijacking in 2009. Since UAVs rely heavily on on-board autopilot to function, it is important to develop an autopilot system that is robust to possible cyber attacks. In this work, we present a biometric system to encrypt the UAV communication by generating a key which is derived from Beta component of the EEG signal of a user. We have developed a safety mechanism that would be activated in case the communication of the UAV from the ground control station gets attacked. This system has been validated on a commercial UAV under malicious attack conditions during which we implement a procedure where the UAV return safely to a `home' position.
\end{abstract}

\section{Introduction}


The role of unmanned aerial vehicles (UAVs) in civilian airspace has been growing, ranging from public safety applications, to commercial use, to personal use by hobbyists. It could be well explained by the increasing affordability of the technology by hobbyists and enthusiasts, which allows them for the creation of innovative applications for the UAV's. This have subsequently led to the occurrence of several severe incidents of different type of attacks on both military and civil UAVs. Several security issues have been demonstrated in recent investigations of cheap consumer UAVs, revealing these systems to be vulnerable to attack.

Commercial activities such as Google's ``Project Wing"~\cite{googlewing}, which has successfully tested its drones for food delivery, and Amazon's ``Prime Air" service~\cite{primeair}, which aims to provide same-day package delivery, would place several drones in commercial airspace, near population centers. This increases with the number of UAVs in civilian airspace and their proximity to people. This increases the potential for, and interest in, potential cyberattacks on those UAVs. These potential threats need to be addressed in order to ensure that a UAV completes its mission and is not used for a malicious purpose. 

In this work, we propose a technique which secures the UAV communication to the ground control station using an encryption key generated using features of a person's electroencephalogram (EEG) signal. UAVs in modern times communicate with each other using small mobile modules called XBee. XBee's provides the functionality of securing the communication using AES encryption standard. We generate an AES encryption key derived from the an EEG signal.  We have developed a demonstration safety mechanism which becomes activated in case there is a detection of potential attack from a third party. This secures UAV communication using a biometric signal. This entire system is validated on a commercially available UAV.

We performed the testing on a UAV, where we encrypt its communication to the ground control station by configuring the XBee's AES encryption key using an EEG biometric key. After configuring the Xbees, we create a simple attack scenario hack, in which the third party or attacker is aware of the key and tries to attack the communication from the UAV to the ground control station. We test our proposed safety solution that enables the UAV to detect that an attack has been attempted and should return back to the `home' station.


\section{Related Work}


There have been several known incidents where civilian and military robots have been remotely compromised for the purposes of taking control of the UAV or making it crash-land. The first most popular known attack on a UAV occurred in 2009, where the Iraqi militants used ``SkyGabber" software to intercept live video feeds from an unsecured communication link used by a Predator Drone~\cite{gorman09wsj}. In October 2011, a key-logging malware was found in the Predator and Reaper ground control stations, likely installed using a removable hard drive. The virus got spreaded to both classified and unclassified computers~\cite{nguyen11virus}.

A more troubling incident that grabbed international attention was the claimed theft of a Sentinel RQ-170 UAV by Iranian forces in December 2012. Hostile agents were able to compromise the control system of the craft and remotely land the UAV, obtaining sensitive data including mission and maintenance data. There are competing theories regarding how the RQ-170 Sentinel may have been lost. The simplest theory is that the loss of the UAV was a result of a technical malfunction, causing the UAV to mistakenly land in Iranian territory~\cite{hartmann2013vulnerability}.

However, a more nefarious possibility is that through a vulnerability in a sensor system, the UAV's GPS could have been intentionally fooled into landing to a location where a hostile agent intended. This type of attack is generally referred to as a ``GPS-Spoofing" attack~\cite{hartmann2013vulnerability, lorenzo2012drone}. An example of this type of attack was demonstrated by a University of Texas at Austin research team, partnered with the Department of Homeland Security to demonstrate the ability to hijack a military UAV. Using relatively inexpensive equipment, these researchers were able to spoof the global position system (GPS) and take complete control of the UAV~\cite{nguyen12college}~\cite{paganini2013hacking}.

Interesting research regarding control security for UAVs is being pursued. A team from the University of Virginia (UVA) and the Georgia Tech Research Institute, operating with the Federal Aviation Administration (FAA), conducted flight tests that evaluated a new class of cyber security solutions on a UAV performing a video surveillance mission. Their goal was to protect computer-controlled remote systems from cyber attacks. It included a new cybersecurity layer called as System-Aware which represents a class of solutions that depends on detailed knowledge of the design of the system being protected. This layer of security provides both complement network and perimeter security solutions and protects against supply chain and insider attacks that may be embedded within a system ~\cite{horowitz2016cybersecurity}.

Most UAV systems are moving their infrastructure towards more network-centric command and control, where all of the components are interconnected through sophisticated mesh networks~\cite{diamond2007application}. This enables fast communication and constant environmental and asset awareness, but introduces security drawbacks. Some military UAV systems, such as the Global Hawk, already employ this type of infrastructure. Public safety and disaster management UAVs are also moving to a similar network architecture for planning and communication~\cite{kuntze2012seneka}. When the components of the system are interconnected through such a network, a compromise of one component can cause a propagation of failures or malicious behavior can occur throughout the whole system.

Other research from UVA and the MITRE Corporation at Creech Air Force Base in Nevada designed and conducted a set of tabletop simulation-based experiments with active military UAV pilots. The purpose of these simulations was to determine the best course of action if a cyberattack was detected and if autonomous behavior could provide a secure and safe solution to a potential attack. The pilots presumed that a System-Aware solution could automatically detect cyber attacks. The pilots were asked to suggest how to control the UAV to restore normal operations. Possibilities included navigating to an earlier waypoint or switching from GPS-based navigation to less accurate, but more trusted, inertial navigation~\cite{horowitz2016cybersecurity}.

An interesting perspective which attempts to solve the cyber-security aspect of UAVs considers the whole scenario of vendor and an attacker as a zero-sum network interdiction game. It is represented as a game where the vendor, also assumed an evader, seeks to choose the optimal path strategy for its UAV, from a source location to a destination location, to evade attacks along the way and minimize its expected delivery time. On the other hand, the attacker or interdictor, aims at choosing the optimal attack locations along the paths traversed by the UAV to interdict the UAV, causing cyber or physical damage, with the goal of maximizing the travel time. Later on it is shown that this  network interdiction game is equivalent to a zero-sum matrix game whose Nash equilibrium (NE) can be derived by solving two linear programming (LP) problems. Solving the LP's would give the expected delivery time under different conditions~\cite{anibalieee17}.

One potential solution is to use biometric information to secure communication between a UAV and its command and control station. This would allow the UAV to verify that its stated operator is issuing the commands to the UAV. To the best of our knowledge, biometric UAV authentication has been limited to facial recognition alone. Facial authentication is problematic since it can be easily deceived by an attacker if they have a picture of the actual operator~\cite{anjos2011counter}. In this way, a more secure biometric feature is needed. We propose to use EEG signal characteristics to secure communication between an operator and a UAV.

\section{Wireless Communication With a UAV}



Communication between a ground station and a civilian UAV is typically done through Zigbee or XBee. ZigBee is based on the international standard 802.15.4. To extend the transmission range, ZigBee is adding mesh networking functionality on top of the 802.15.4 standard, in which single messages are forwarded through the network to its destination node. Depending on the frequency band used, transmission rates can vary. Typically it ranges from 20kbit/s to 250kbit/s~\cite{evans2003bzzzz}.

Zigbee uses IEEE 802.15.4 protocol as its MAC layer. IEEE 802.15.4 sets the encryption algorithm to use when cyphering the data to transmit. However, the standard does not specify how the keys have to be managed or what kind of authentication policies have to be applied. These issues are treated in the upper layers which are managed by ZigBee. The encryption algorithm used is AES (Advanced Encryption Standard) with a 128b key length (16 Bytes). The AES algorithm is not only used to encrypt the information but also validates the data which is sent~\cite{lu2002integrated}. This Data Integrity is achieved using a Message Integrity Code (MIC) also named as Message Authentication Code (MAC) which is added to the message. So, a message is received from a non-trusted node we will see that the MAC generated for the sent message does not correspond to the one what would be generated using the message with the current secret Key, so we can discard this message. The MAC can have different sizes: 32, 64, 128 bits, however it is always created using the 128b AES algorithm. The MAC's size is just the bit length which is attached to each frame. Data security is accomplished by encrypting the data payload field with the 128b Key.  ZigBee implements two extra security layers on top of the 802.15.4 layer: Network and Application security layers. All three security policies rely on the AES 128b encryption algorithm. There are three kinds of keys: 

\noindent \textbf{Master key:}
These are pre-installed in each node. Their function is to keep confidential the Link Keys exchange between two nodes in the Key Establishment Procedure (SKKE)~\cite{yuksel2008zigbee}.

\noindent \textbf{Link Keys:}
These are unique between each pair of nodes. These keys are managed by the Application level. They are used to encrypt all the information between each two devices, for this reason more memory resources are needed in each device
~\cite{yuksel2008zigbee}.

\noindent \textbf{Network Keys:}
These are a unique 128b key shared among all the devices in the network. It is generated by the Trust Center and regenerated at different intervals. Each node has to get the Network Key in order to join the network. Once the trust center decides to change the Network Key, the new one is spread through the network using the old Network Key (see image above about ``ZigBee Residential Mode"). Once this new key is updated in a device, its Frame Counter (see in the previous sections)
is initialized to zero. This Trust Center is normally the Coordinator, however it can be a dedicated device. It has to authenticate and validate each device which attempts to join the network.

\section{Approach}

\begin{figure}[t]
\centering
\includegraphics[width=1\columnwidth]{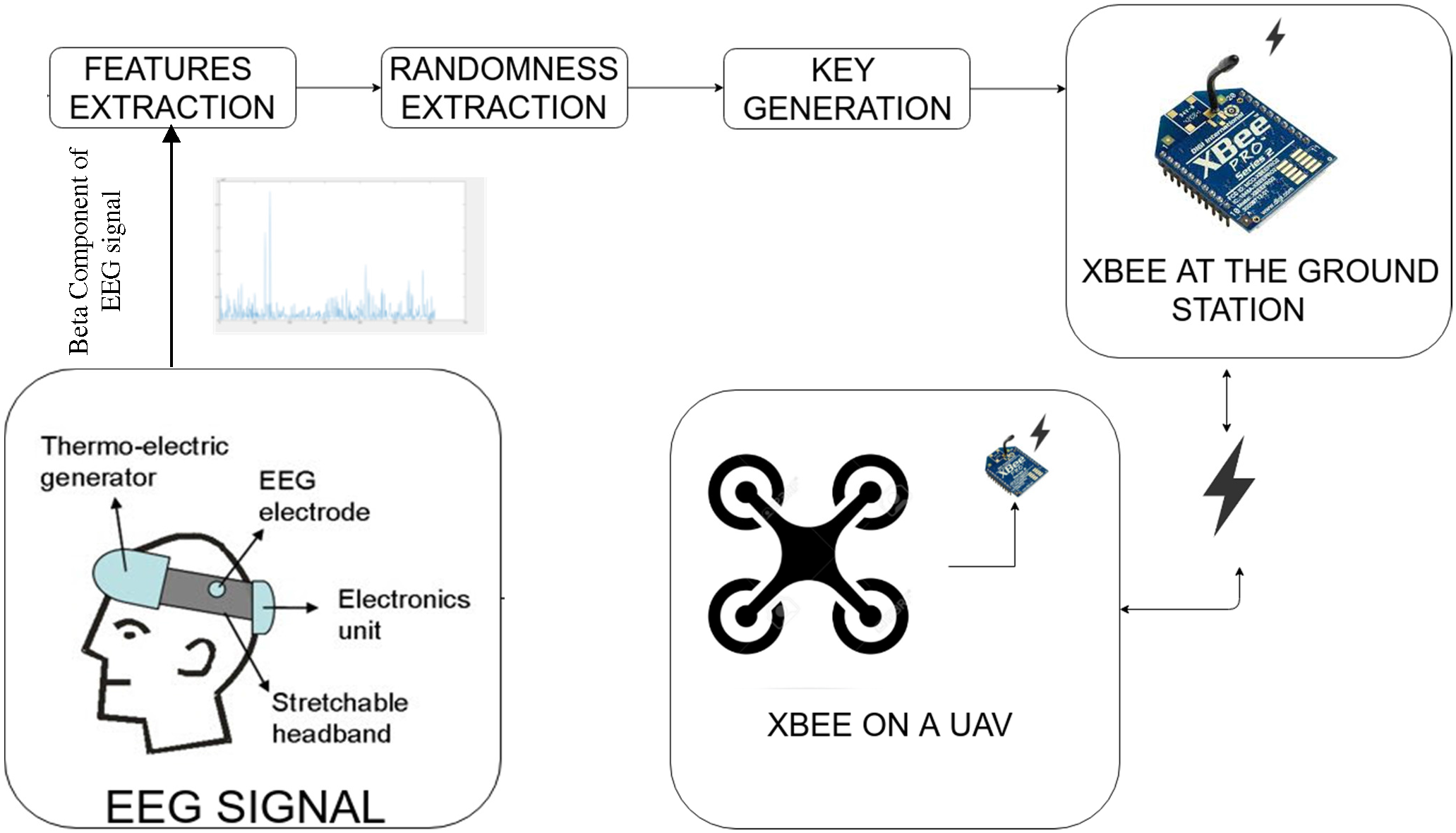} 
\caption{Basic block diagram of the system overview.}  
\label{fig:eeg-xbee}
\vspace{-10pt}
\end{figure}

The EEG signal is unique; to a person and values overtime. It is possible to generate a key unique to a particular user. Also, based on different user activity and different state of mind, the EEG signal of even a same person will be different.
Moreover, this unique signal changes every few hours at different state of mind, which means that it cannot be permanently ``stolen.'' This unique key can be used for encrypting AES data like what is used in Zigbee communication. We have developed a robust method for utilizing brain EEG signal characteristics to generate the cryptographic key for AES data encryption and decryption. In this section, we describe our method for securing a UAV communication using this EEG signal. We configure the AES encryption key of the XBee of both the UAV and the ground control station with the Key generated from the above procedure. We also implement a safety backtrack path procedure in case the communication is attacked.

\subsection{EEG Signal Properties}

We obtain a user's EEG signal recorded using the Mindwave EEG sensor~\cite{salabun2014processing}. This device safely measures and outputs the EEG power spectra (alpha waves, beta waves, etc), NeuroSky eSense meters (attention and meditation), and eye blinks. The device consists of a headset, an ear-clip, and a sensor arm. The headset’s reference and ground electrodes are on the ear clip and the EEG electrode is on the sensor arm, resting on the forehead above the eye (see Figure \ref{fig:eeg-xbee}). It operates using battery power.

We chose to use Beta waves from the EEG signal as the basis for our analysis. Beta waves are in the frequency range of 12 and 30 Hz, but are often divided into $\beta1$ (low Beta) and $\beta2$ (high Beta) to get a more specific range. The waves are small and fast, associated with focused concentration and best defined in central and frontal areas. There is an increase of $\beta$ activity when a person concentrates on tasks such as resisting or suppressing movement or solving a math task.

\begin{figure}[h]
\centering
\includegraphics[width=0.43\textwidth]{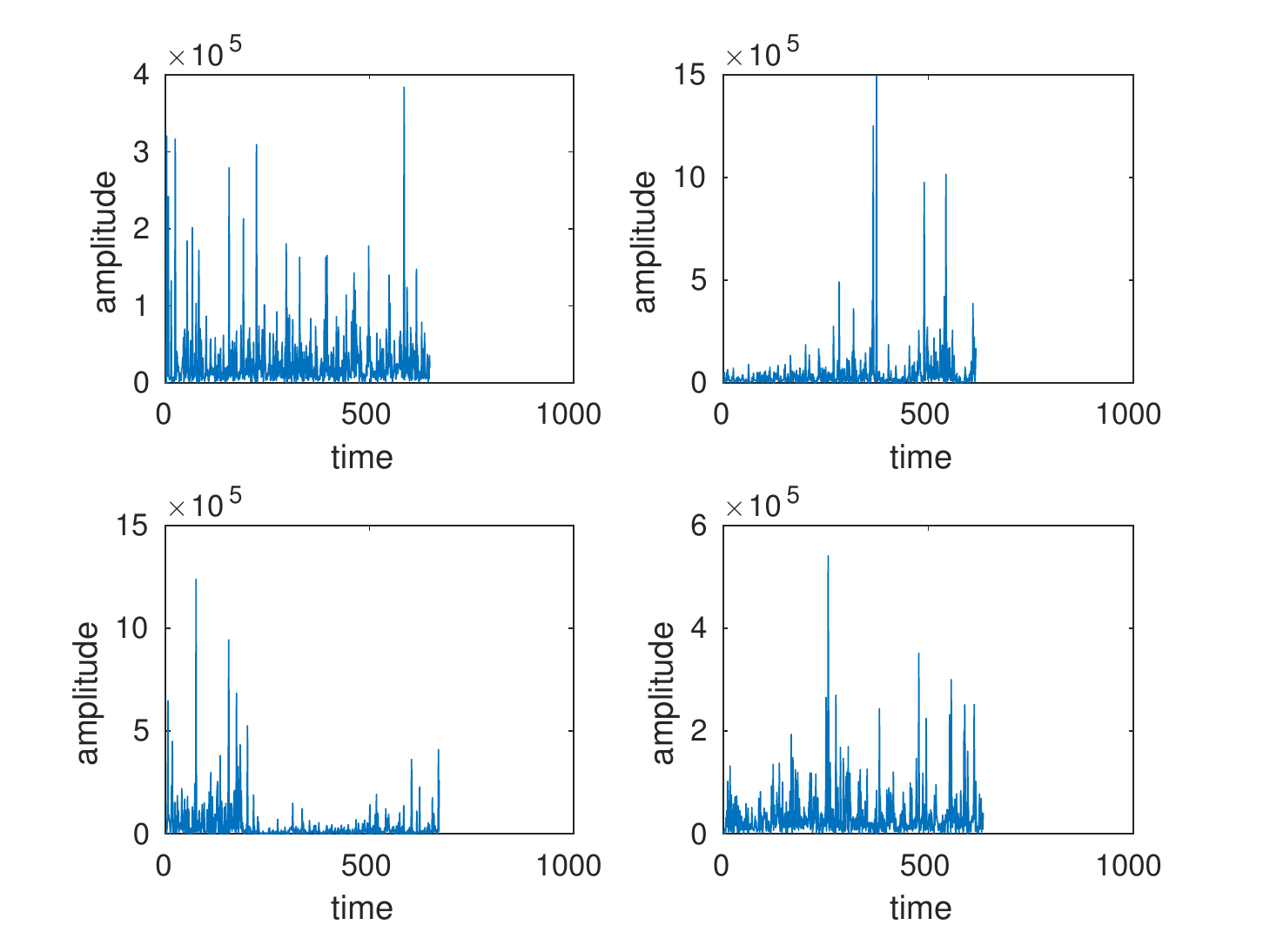} 
\caption{Beta component of EEG waveform of 4 different people. The patterns in the $\beta$ waves are unique to each individual, making them ideal for biometric encryption.}  
\label{fig:beta}
\vspace{-10pt}
\end{figure}




\subsection{Feature Extraction}

We record a EEG signal (Beta waves) from a specific user for time period $T$. The Beta waves are amplified by amplifier value $A$ and matched via high order Legendre Polynomials, where Legendre Differential equation is given by:
\begin{equation}
\frac{\partial}{\partial x}[(1-x^{2})\frac{\partial }{\partial x} p_{n}\left ( x \right )] + n(n+1)p_{n}(x)=0.
\end{equation}

Legendre polynomials are computed using Rodrigues's formula, which is given by:
\begin{equation}
p_{n}(x)=\frac{1}{2^{n}n!}\frac{\partial^n}{\partial x^n}[(x^2 -1)^n].
\end{equation}

The $n$-degree equation used for fitting data is given by:
\begin{equation}
y\left ( x \right )=a_{0}+\sum_{1}^{n}a_{i}p_{i\left ( x \right )}.
\end{equation}

The polynomial coefficients $a_{0},a_{1},...a_{n}$ are combined together with the time window of size $T$ and the amplitude multiplier $A$ to form the raw feature vector $z:=\left \{ ca_{0},ca_{1},ca_{2},...ca_{n},A,T \right \}$ where $c$ is a constant to magnify the difference between coefficients. We map $z$ to $w$ such that $w=z \times M+\gamma $ where $M$ is an $n \times n$ invertible matrix which satisfies $\sum _{i}m_{i,j}=1$; $\gamma$ is a random vector whose elements are within the range $\left [ 2^{-\theta },2^{\theta } \right ]$.

The polynomial coefficients are combined together with the time window
size $T$ and the amplitude multiplier $A$ to form the raw feature vector. Since attackers can reconstruct the original EEG waveform given the feature vector, we map the feature vector with some random vector using Linear transformation. This results as a random feature vector $w$~\cite{huang2016robust}.

\subsection{Randomness Extraction}

After obtaining feature vector $w$, we use a reusable fuzzy extractor constructed from $(n,k)$-BCH codes (The BCH codes form a class of cyclic error-correcting codes to correct errors occurred~\cite{huang2016robust}.) with generator function to extract enough randomness from it. Randomness provides the functionality of representing the feature vector in different form so that attacker cannot reconstruct the original signal.

The randomness extracted from each feature $r_{i}$ is computed as $r_{i}=H_{x}(w_{i})$, where $H_{x}$ is a hash function in a universal hash family. The universal hash family $H$ is a class of hash functions. $H$ is defined to be universal if the possibility of a pair of distinct keys being mapped into the same index is less than $1/l$($l$ is the length of the randomness string). The hashing operation is performed after making a random choice of hash function from the universal class $H$. The universal hash function already gives the optimal length of extracted randomness~\cite{huang2016robust}.

We also compute the syndrome $S_{c}$ of feature values for future authentication. If the feature element is viewed as $w_{i}(x)=w_{i_{0}}+w_{i_{1}}x+...+w_{i_{n-1}}x^{n-1}$, every element $w_{i}$ has a corresponding syndrome $S_{c_{i}}$ for $(n,k)$-BCH codes:
\begin{equation}
S_{c_{i}}=w_{i}(x) \text{mod} g(x)= \left \{ w_{i}(\alpha^{1}),w_{i}(\alpha^{2}),...,w_{i}(\alpha^{2t}) \right \}.
\end{equation}

\subsection{Key Generation}

Next, we generate the key based on the features and pre-program the specific 
UAV with that key to secure the communication channel. This is a practical way to ensure that both the ground control station Xbee and the XBee on-board UAV obtain exactly the same key for encryption and decryption. The key $K$ is generated based on chosen extracted randomness from the previous step~\cite{huang2016robust}. The key generation technique is given below.

We randomly choose $q$ constants  $1 \leq j_{1}\leq ...\leq j_{q}\leq n$ to pick up several features and produces a permuted feature vector $v:= \left \{  w_{j_{1}},...,w_{j_{q}}\right \}$.

The key $K$ generated is based on chosen random extracted randomness $r_{j_{i}}:K:=r_{j_{1}}|| ... || r_{j_{q}},$ where $ || $ denotes concatenation.

\subsection{Configuring XBee with the Key Generated}

After generating the key using the above procedure we configure the XBee's AES encryption key parameter to use the generated key for communication. For this experiment, we used the Mindwave sensor and Alienware 15' with i7 6820 HK  processor to create the EEG system.
The architecture of the EEG system is described in Figure~\ref{fig:eeg-xbee}.

We utilized a commercially available UAV to conduct this experiment. The UAV uses the Pixhawk as its controller and had ardroid as the CPU which had the XBee connected to in roder to communicate with the ground control station.
The UAV and the base station were wirelessly connected using Xbee transmitter and receivers.

After configuring the AES encryption key of the XBee with the generated encryption key, we tested the communication of UAV with the Xbee connected to the ground control station. The AES key configuration ensured secured communication of the UAV to the ground control station. However, we have also introduced a scenario where an attacker is trying to intercept the communication between the UAV and ground control station for the primary purpose of controlling the UAV for its own purpose. For simplicity, we have assumed that the attacker already knows the key generated and has configured its own device with that key and to maliciously communicate with the UAV.

\begin{figure}[h]
\centering
\includegraphics[width=0.48\textwidth]{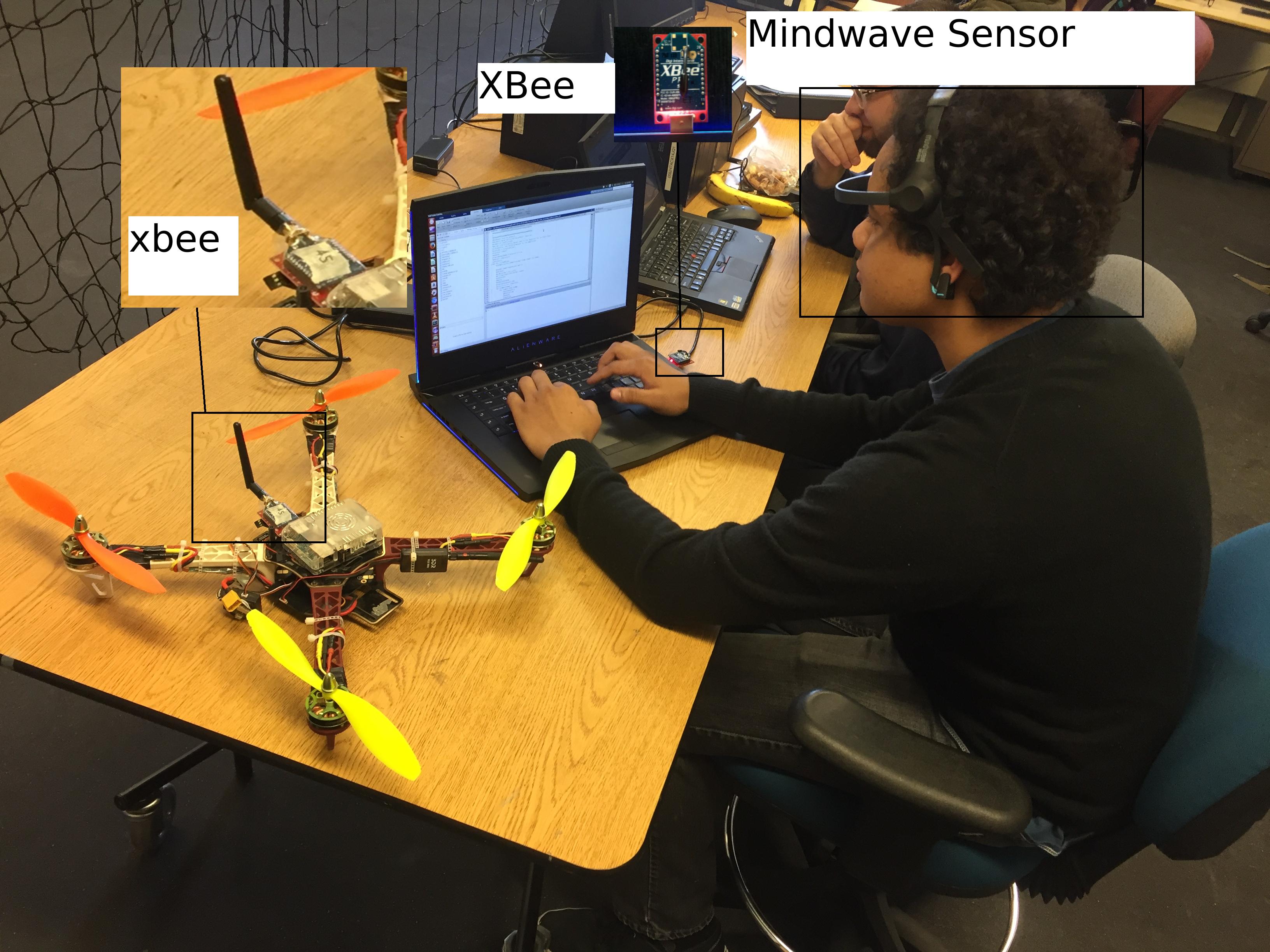} 
\caption{Experimental Setup.}  
\label{fig:expsetup}
\end{figure}

As a safety measure, we have preconfigured the UAV's Xbee to receive the commands from the ground control station Xbee's address. If the attacker tries to send the control signals from it's device then from the attacker's packet address we  verify that a third party is intervening and we activate the Return-To-Launch control signal in the UAV. This would mean that the UAV identified that an attack was attempted and should return to its starting location. The RTL (Return-To-Launch mode) aids the UAV navigation from its current position to hover above the home position. RTL is a GPS-dependent move, so it is essential that GPS lock is enabled before attempting to use this mode. The algorithm is described below as Algorithm 1.

\begin{algorithm}
\caption{RTL mode activation in UAV}
\begin{algorithmic} 
\STATE $getAddress\gets xbeedata.getAddress()$
\IF {$getAddress\neq groundcontrolstation.getAdress()$} 
        \STATE $LockGPS()$
        \STATE $ReturnToLaunch()$
\ELSE
        \STATE $Continue;$ 
\ENDIF 
\end{algorithmic}
\end{algorithm}
\vspace{-10pt}

The LockGPS() function ensures that the sensor is not affected by any other  way since it becomes completely independent of the rest of the communication process.
 
We also propose another methodology where, in case a hack is attempted, the Xbee sends predefined signal to the ground control station which signals the station to configure the XBees (both at the ground control station and at the UAV) with a new key. We then run at the ground control station the same pipeline of key generation from the EEG signal and generate another key to ensure the communication is secure and configure both the XBees.


We describe the algorithm below (Algorithm \ref{key change}):
\begin{algorithm}
\caption{Key Change request in UAV}
\begin{algorithmic} 
\STATE $getAddress\gets xbeedata.getAddress()$
\IF {$getAddress\neq groundcontrolstation.getAdress()$} 
        \STATE $LockGPS()$
        \STATE $SendKeyChangeToGroundControlStation()$
        \STATE $WaitForKey()$
\ELSE
        \STATE $Continue();$ 
\ENDIF 
\end{algorithmic}
\label{key change}
\end{algorithm} 
\vspace{-10pt}

Another attempt to ensure a secure communication is to regularly change the key generated and configure the Xbees at regular intervals of time. This way we achieve quite robust and secure way of communication in the UAVs.

\section{Results}

In the initial setup we collect the EEG data and activated our key generation pipeline to generate a key. The data were collected from a user performing a specific task which involves activating the Beta component of the EEG signal. The collected data (around 1000 data points), are fed to the data to our key generation pipeline that involves Features extraction, Randomness extraction and Key generation. 

Then we configure the XBee's in AT mode to ensure that XBee's AES encryption mode is enabled and uses the Key generated from our pipeline. Since the EEG data is inconsistent even for the same user at different time intervals, the generated Key from our pipeline would be different, thus ensuring uniqueness of the Key generated. This enables users to configure the XBee's AES key to different values. This setup is shown in Figure~\ref{fig:expsetup}.

A normal EEG waveform of a single person is shown in Figure~\ref{fig:waveform}:
\begin{figure}
\centering
\includegraphics[width=0.48\textwidth]{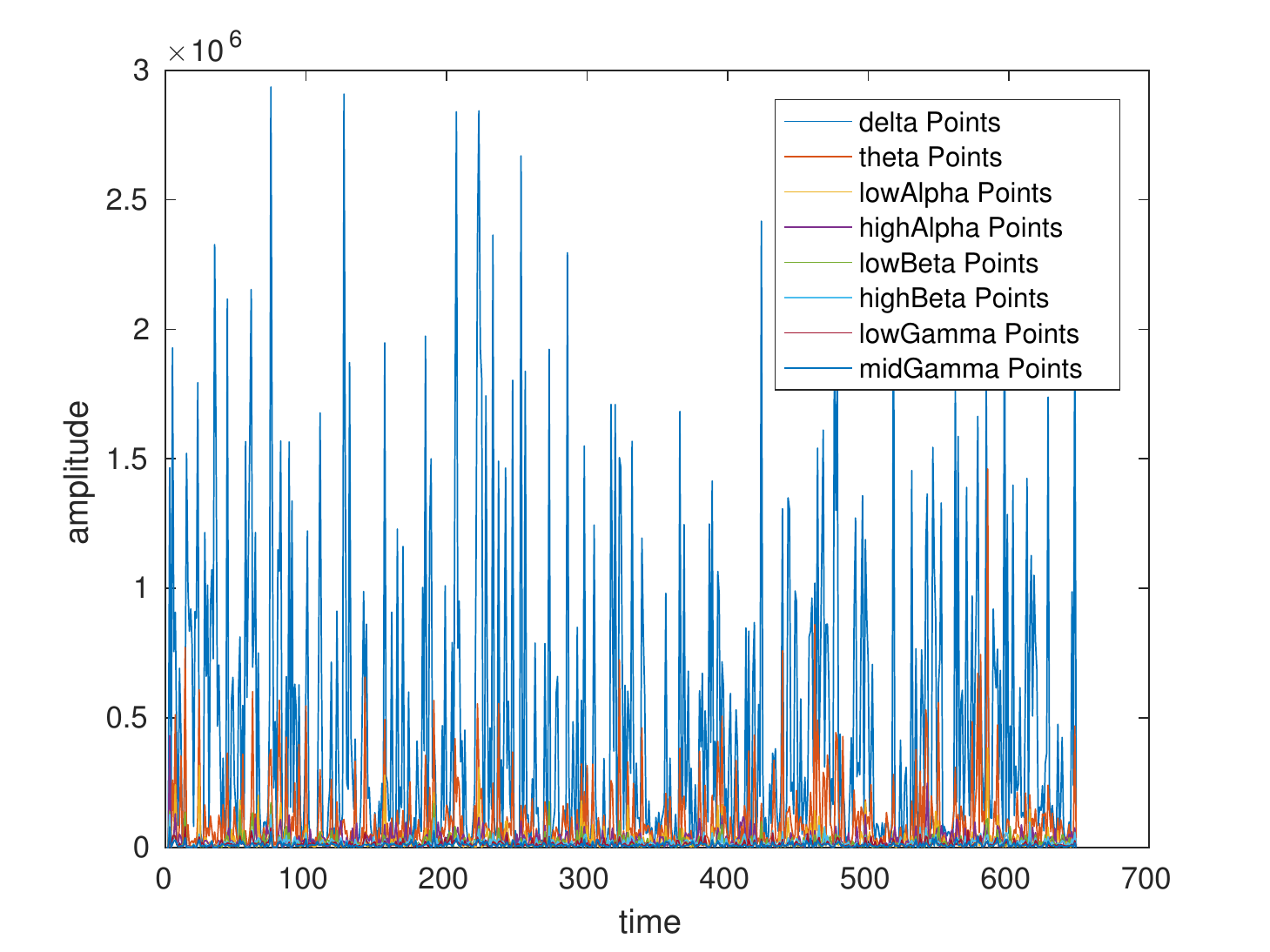} 
\caption{Sample EEG waveform(with all the components) of a user performing a specific mental task.}  
\label{fig:waveform}
\vspace{-10pt}
\end{figure}
We extract the Beta components of different people which was collected under similar tasks. It is shown in Figure \ref{fig:waveform}.

Xbee has 2 basic modes of interaction: AT and AP2 mode. AT mode is also referred as ``Transparent" mode.  In AT mode, any data sent to the XBee module is immediately sent to the remote module identified by the Destination Address in memory.  When the module is in AT mode, it can be configured by the user or a host microcontroller by first placing the module in Command mode and then sending predefined AT commands through the UART port.  This mode is useful when you don't need to change destination addresses very often, or you have a very simple network, or simple point to point communication
Since Xbee's can be configured only in AT mode we had to stop the ongoing communication operation which was going on in AP2 mode. In AT mode no communication takes place so even a potential attack would fail to communicate. Xbee's inbuild encryption also gets disabled in AT mode, so it is important to give proper attention while configuring XBee's.

\begin{figure}
\centering
\includegraphics[width=0.48\textwidth]{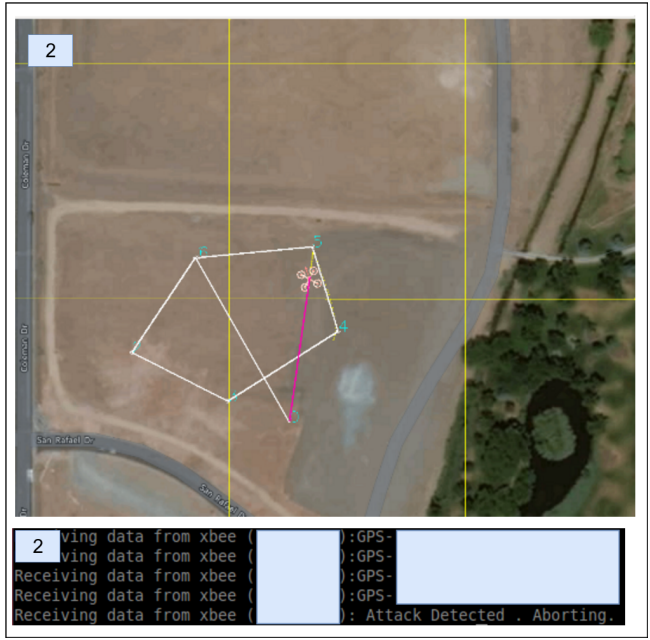} 
\caption{Waypoints set for the experiment in the second configuration. The attack was discovered after the UAV navigated from waypoint 5 and Return-to-Launch (RTL) was enabled.}  
\label{fig:UAV_p2}
\vspace{-10pt}
\end{figure}

We performed our proposed safety mechanism using an assembled quadcopter with an onboard autopilot and the Xbee's to communicate with ground control station. We set up the waypoints for the UAV using mission planner software.  For our experiment we set up different waypoints at different configurations and tested our methods at different times. It is shown from Figure ~\ref{fig:UAV_p1} to Figure ~\ref{fig:UAV_p2}.
\begin{figure}
\centering
\includegraphics[width=0.48\textwidth]{HACK1.png} 
\caption{Waypoints set for the experiment in the first configuration. The attack was discovered after the UAV navigated from waypoint 3 and Return-to-Launch (RTL) was enabled.}  
\label{fig:UAV_p1}
\vspace{-10pt}
\end{figure}

%

 The purpose was to travel these way-points and return to the base location in case an attack is detected. We introduced a third party attacking mechanism, and for simplistic purpose we made the third party aware of the key generated which the UAV's Xbee is using to communicate with the ground control station. As the third party starts attacking and maliciously sending control signals to the UAV, our algorithm successfully detects the intervention (since the received packets at the UAV's XBee has different source address).  After detection of the intervention, the UAV initiates its RTL mechanism and return to the base GPS location without completing the directed trajectory. 

We tested our other approach of changing the key when an attack is detected. During this test we setup the same waypoints  and introduced a similar type of attack along the way. After successful detection of the intervention, the algorithm sent a key change request to the ground control station, during which, the UAV's communication is restricted to the ground control station and it hovers at a specified location where the attack was attempted. After the Xbee is configured to a new AES key, the navigation is resumed to the destined location.

\section{Conclusion}
We have provided an approach for biometric encryption of a UAV communicating with the ground control station. We have also  provided a safety mechanism for the UAV in case a third-party intervention is detected along the way. This approach can be used for any UAV scenario where cyberattacks are a particular concern. Our approach not only adds a layer of additional security to the UAV but also provides a unique way for securing the UAV with low-cost resources.  

In the future work, we plan to further extend our authentication scheme to multi-UAV scenarios \cite{La_RAS_2012, La_TCST_2015}, where a cluster of UAVs aim to authenticate their controller. A possible approach is to have each member in the all UAVs (a cluster) sequentially verify the controller one by one utilizing the proposed authentication scheme.  Formation control and cooperative learning in multi-robot systems can be utilized to enhance the safety security mechanism \cite{La_SMCA_2015, La_SMCB_2013}.

\section*{Acknowledgements}

This material is based upon work supported by the National Aeronautics and Space Administration (NASA)  under Grant No. NNX10AN23H issued through the Nevada NASA Space Grant, and Grant No. NNX15AI02H issued through the Nevada NASA Research Infrastructure Development Seed Grant.

\bibliographystyle{abbrv}
\bibliography{eeg}

\begin{thebibliography}{10}

\bibitem{paganini2013hacking}
Hacking drones ... overview of the main threats.

\bibitem{primeair}
Amazon.
\newblock Amazon prime air, 2016.

\bibitem{anibalieee17}
W.~S. Anibal~Sanjab and T.~Başar.
\newblock Prospect theory for enhanced cyber-physical security of drone
  delivery systems: A network interdiction game.
\newblock In {\em Proc. of the IEEE International Conference on Communications
  (ICC), Communication and Information Systems Security Symposium, Paris,
  France,}, 2017.

\bibitem{anjos2011counter}
A.~Anjos and S.~Marcel.
\newblock Counter-measures to photo attacks in face recognition: a public
  database and a baseline.
\newblock In {\em Biometrics (IJCB), 2011 international joint conference on},
  pages 1--7. IEEE, 2011.

\bibitem{diamond2007application}
S.~M. Diamond and M.~G. Ceruti.
\newblock Application of wireless sensor network to military information
  integration.
\newblock In {\em Industrial Informatics, 2007 5th IEEE International
  Conference on}, volume~1, pages 317--322. IEEE, 2007.

\bibitem{evans2003bzzzz}
C.~Evans-Pughe.
\newblock Bzzzz zzz [zigbee wireless standard].
\newblock {\em IEE review}, 49(3):28--31, 2003.

\bibitem{lorenzo2012drone}
L.~Franceschi-Bicchierai.
\newblock Drone hijacking? that’s just the start of gps troubles, July 2012.

\bibitem{gorman09wsj}
S.~Gorman.
\newblock Insurgents hack u.s. drones, 2009.

\bibitem{hartmann2013vulnerability}
K.~Hartmann and C.~Steup.
\newblock The vulnerability of uavs to cyber attacks-an approach to the risk
  assessment.
\newblock In {\em Cyber Conflict (CyCon), 2013 5th International Conference
  on}, pages 1--23. IEEE, 2013.

\bibitem{horowitz2016cybersecurity}
B.~M. Horowitz.
\newblock Cybersecurity for unmanned aerial vehicle missions., April 2016.

\bibitem{huang2016robust}
P.~Huang, B.~Li, L.~Guo, Z.~Jin, and Y.~Chen.
\newblock A robust and reusable ecg-based authentication and data encryption
  scheme for ehealth systems.
\newblock In {\em Global Communications Conference (GLOBECOM), 2016 IEEE},
  pages 1--6. IEEE, 2016.

\bibitem{kuntze2012seneka}
H.-B. Kuntze, C.~W. Frey, I.~Tchouchenkov, B.~Staehle, E.~Rome, K.~Pfeiffer,
  A.~Wenzel, and J.~W{\"o}llenstein.
\newblock Seneka-sensor network with mobile robots for disaster management.
\newblock In {\em Homeland Security (HST), 2012 IEEE Conference on Technologies
  for}, pages 406--410. IEEE, 2012.

\bibitem{La_TCST_2015}
H.~M. La, R.~Lim, and W.~Sheng.
\newblock Multirobot cooperative learning for predator avoidance.
\newblock {\em IEEE Transactions on Control Systems Technology}, 23(1):52--63,
  Jan 2015.

\bibitem{La_RAS_2012}
H.~M. La and W.~Sheng.
\newblock Dynamic target tracking and observing in a mobile sensor network.
\newblock {\em Robotics and Autonomous Systems}, 60(7):996 -- 1009, 2012.

\bibitem{La_SMCB_2013}
H.~M. La and W.~Sheng.
\newblock Distributed sensor fusion for scalar field mapping using mobile
  sensor networks.
\newblock {\em IEEE Transactions on Cybernetics}, 43(2):766--778, April 2013.

\bibitem{La_SMCA_2015}
H.~M. La, W.~Sheng, and J.~Chen.
\newblock Cooperative and active sensing in mobile sensor networks for scalar
  field mapping.
\newblock {\em IEEE Transactions on Systems, Man, and Cybernetics: Systems},
  45(1):1--12, Jan 2015.

\bibitem{lu2002integrated}
C.-C. Lu and S.-Y. Tseng.
\newblock Integrated design of aes (advanced encryption standard) encrypter and
  decrypter.
\newblock In {\em Application-Specific Systems, Architectures and Processors,
  2002. Proceedings. The IEEE International Conference on}, pages 277--285.
  IEEE, 2002.

\bibitem{googlewing}
M.~McFarland.
\newblock {Google drones will deliver Chipotle burritos at Virginia Tech},
  September 2016.

\bibitem{nguyen11virus}
T.~C. Nguyen.
\newblock Virus attacks military drones, exposes vulnerabilities, October 2011.
\newblock Retrieved 6/7/13.

\bibitem{nguyen12college}
T.~C. Nguyen.
\newblock How college students hijacked a government spy drone., 2012.
\newblock Retrieved 6/7/13.

\bibitem{salabun2014processing}
W.~Sa{\l}abun.
\newblock Processing and spectral analysis of the raw eeg signal from the
  mindwave.
\newblock {\em Przeglad Elektrotechniczny}, 90(2):169--174, 2014.

\bibitem{yuksel2008zigbee}
E.~Y{\"u}ksel, H.~R. Nielson, and F.~Nielson.
\newblock Zigbee-2007 security essentials.
\newblock In {\em Proc. 13th Nordic Workshop on Secure IT-systems}, pages
  65--82, 2008.

\end{thebibliography}

\end{document}